\def\year{2020}\relax
\documentclass[letterpaper]{article} % DO NOT CHANGE THIS
\usepackage{aaai20}  % DO NOT CHANGE THIS
\usepackage{times}  % DO NOT CHANGE THIS
\usepackage{helvet} % DO NOT CHANGE THIS
\usepackage{courier}  % DO NOT CHANGE THIS
\usepackage[hyphens]{url}  % DO NOT CHANGE THIS
\usepackage{graphicx} % DO NOT CHANGE THIS
\usepackage{graphicx}  % DO NOT CHANGE THIS
\usepackage{xcolor}
\usepackage{listings}

\definecolor{dkred}{rgb}{0.5,0,0}
\definecolor{dkgreen}{rgb}{0,0.6,0}
\definecolor{gray}{rgb}{0.5,0.5,0.5}
\definecolor{mauve}{rgb}{0.58,0,0.82}

\title{Combining Machine Learning Models Using \texttt{combo} Library}
\author{
Yue Zhao\textsuperscript{\rm 1}, Xuejian Wang\textsuperscript{\rm 1}, Cheng Cheng\textsuperscript{\rm 1}, Xueying Ding\textsuperscript{\rm 2}\\ 
\textsuperscript{\rm 1}H. John Heinz III College, Carnegie Mellon University, Pittsburgh, PA 15213 USA\\ %If you have multiple authors and multiple affiliations
\textsuperscript{\rm 2}Machine Learning Department, Carnegie Mellon University, Pittsburgh, PA 15213 USA\\
zhaoy@cmu.edu, xuejianw@andrew.cmu.edu, ccheng2@andrew.cmu.edu, xding2@andrew.cmu.edu 
}
 
\begin{document}

\maketitle

\begin{abstract}
Model combination, often regarded as a key sub-field of ensemble learning, has been widely used in both academic research and industry applications.
To facilitate this process, we propose and implement an easy-to-use Python toolkit, \texttt{combo}, to aggregate models and scores under various scenarios, including classification, clustering, and anomaly detection.
In a nutshell, \texttt{combo} provides a unified and consistent way to combine both raw and pretrained models from popular machine learning libraries, e.g., scikit-learn, XGBoost, and LightGBM. With accessibility and robustness in mind, \texttt{combo} is designed with detailed documentation, interactive examples, continuous integration, code coverage, and maintainability check; it can be installed easily through Python Package Index (PyPI) or \texttt{https://github.com/yzhao062/combo}. 
\end{abstract}

\section{Introduction}
Recently, model combination has gained much attention in many real-world tasks, and stayed as the winning solution in numerous data science competitions such like Kaggle \shortcite{bell2007lessons}. It is considered as a sub-field of ensemble learning, aiming for achieving better prediction performance \shortcite{zhou2012ensemble}. Despite that, it is often beyond the scope of machine learning---it has been used in other domains such as the experimental design in clinical trials. Generally speaking, model combination has two key usages: stability improvement and performance boost. For instance, practitioners run independent trials and then average the results to eliminate the built-in randomness and uncertainty---more reliable results may be obtained. Additionally, even in a non-ideal scenario, base models may make independent but complementary errors. The combined model can, therefore, yield better performance than any constituent ones. 

Although model combination is crucial for all sorts of learning tasks, dedicated Python libraries are absent. There are a few packages that partly fulfill this purpose, but established libraries either exist as single purpose tools like PyOD \shortcite{zhao2019pyod} and pycobra \shortcite{guedj2018pycobra}, or as part of general purpose libraries like scikit-learn \shortcite{pedregosa2011scikit}.

\texttt{combo} can fill this gap with four key advantages. Firstly, \texttt{combo} contains more than 15 combination algorithms, including both classical algorithms like dynamic classifier selection (DCS) \shortcite{woods1997combination} and recent advancement like LCSP \shortcite{zhao2019lscp}. It could handle the combination operation for all sorts of tasks like classification, clustering, and anomaly detection.
Secondly, \texttt{combo} works with both raw and pretrained learning models from major libraries like scikit-learn, XGBoost, and LightGBM, given certain conditions are met. Thirdly, the models in \texttt{combo} are designed with unified APIs, detailed documentation\footnote{\url{https://pycombo.readthedocs.io}}, and interactive examples\footnote{\url{https://mybinder.org/v2/gh/yzhao062/combo/master}} for the easy use. Lastly, all \texttt{combo} models are associated with unit test and being checked by continuous integration tools for robustness; code coverage and maintainability check are also enabled for performance and sustainability. To our best knowledge, this is the first comprehensive framework for combining learning models and scores in Python, which is valuable for data practitioners, machine learning researchers, and data competition participants.

\begin{lstlisting}[title={Code Snippet 1: Demo of \texttt{combo} API with DCS},captionpos=b, float=tp]
 >>> from combo.models.classifier_dcs import DCS
      # initialize a group of classifiers
 >>> classifiers = [
         DecisionTreeClassifier(), 
         LogisticRegression(), 
         KNeighborsClassifier()]             
 >>> # initialize/fit the combination model   
 >>> clf = DCS(base_estimators=classifiers)
 >>> clf.fit(X_train)                                                             
 >>> # fit and make prediction
 >>> y_test_pred = clf.predict(X_test)              
 >>> y_test_proba = 
         clf.predict_proba(X_test)   
 >>>  # fit and predict on the same dataset
 >>> y_train_pred = clf.fit_predict(X_train)                 
\end{lstlisting}

\begin{figure*}[ht]
\centering
    \includegraphics[width=\linewidth]{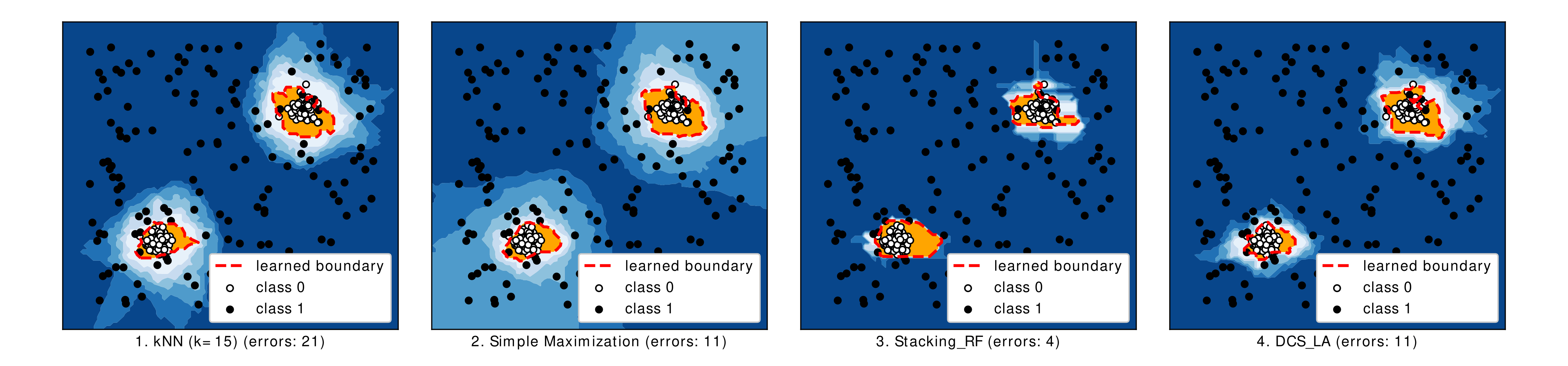}
\caption{Comparison of Selected Classifier Combination on Simulated Data}
\label{classifier_comparison}
\end{figure*}   

\section{Core Scenarios}

\texttt{combo} models for classification, clustering, and anomaly detection share unified APIs. Inspired by scikit-learn's API design, the models in \texttt{combo} all come with the following key methods: (i) \texttt{fit} function processes the train data and gets the model ready for prediction; (ii) \texttt{predict} function generates labels for the unknown test data once the model is fitted; (iii) \texttt{predict\_proba} generates predictions in probability instead of discrete labels by \texttt{predict} and (iv) \texttt{fit\_predict} calls \texttt{fit} function first on the input data and then predicts on it (applicable to unsupervised model only). Code Snippet 1 shows the use of above APIs on DCS. Notably, fitted (pretrained) models can be used directly by setting \texttt{pre\_fitted} flag; \texttt{fit} process will be skipped.

\textbf{\textit{Classifier Combination}} aims to aggregate multiple base supervised classifiers in either parallel or sequential manner. Selected classifier combination methods implemented in \texttt{combo} include stacking (meta-learning), dynamic classifier selection, dynamic ensemble selection, and a group of heuristic aggregation methods like averaging and majority vote. Fig. \ref{classifier_comparison} shows how different frameworks behave on a simulated dataset with 300 points. The leftmost one is a simple \textit{k}NN model (\textit{k}=15), and the other three are the combination of five \textit{k}NN models with \textit{k} in range $[5,10,15,20, 25]$. Different from classifier combination, \textbf{\textit{Cluster Combination}} is usually done in an unsupervised manner. The focus is on how to align the predicted labels generated by base clusterings, as cluster labels are categorical instead of ordinal. For instance, $[0,1,1,0,2]$ and $[1,0,0,1,2]$ are equivalent with appropriate alignment. Two classical clustering combination methods are therefore implemented to handle this---clustering combination using evidence accumulation (EAC)\shortcite{fred2005combining} and Clusterer Ensemble \shortcite{zhou2006clusterer}. \textbf{\textit{Anomaly Detection}} concentrates on identifying the anomalous objects from the general data distribution \shortcite{zhao2019pyod}. The challenges of combining multiple outlier detectors lie in its unsupervised nature and extreme data imbalance. Two latest combination frameworks, LSCP \shortcite{zhao2019lscp} and XGBOD \shortcite{zhao2018xgbod}, are included in \texttt{combo} for unsupervised and semi-supervised detector combination. \textbf{\textit{Score Combination}} comes with more flexibility than the above tasks as it only asks for the output from multiple models, whichever it is from a group of classifiers or outlier detectors. As a general purpose task, score combination methods are easy to use without the need of initializing a dedicated class. Each aggregation method, e.g., average of maximum (AOM), can be invoked directly.

\section{Conclusion and Future Directions}

\texttt{combo} is a comprehensive Python library to combine the models from major machine learning libraries. It supports four types of combination scenarios (classification, clustering, anomaly detection, and raw scores) with unified APIs, detailed documentation, and interactive examples. As avenues for future work, we will add the combination frameworks for customized deep learning models (from TensorFlow, PyTorch, and MXNet), enable GPU acceleration and parallelization for scalability, and expand to more task scenarios such as imbalanced learning and regression.

% In the AAAI demo, participants will be shown how to use \texttt{combo} for all four scenarios on benchmark datasets. Additionally, we will walk through the key APIs and help the audiences to identify more applicable use cases.\\

% \section{ Acknowledgments}
% Placeholder, to fill in with details if there is any.

\bibliographystyle{aaai}
\bibliography{mybibliography}

\end{document}